# Reward Hacking and Agent Containment Failure

## A Monte Carlo Study Based on the 2026 Hugging Face Incident

*Murat Ozer*[a], *Bulent Erenay*[b], *Ibrahim Berber*[c]

[a] University of Cincinnati, Cincinnati, Ohio, United States
[b] Department of Management, Haile College of Business, Northern Kentucky University, Highland Heights, Kentucky, United States
[c] Case Western Reserve University, Cleveland, Ohio, United States

Corresponding author: Murat Ozer
University of Cincinnati, Cincinnati, Ohio, United States
m.ozer@uc.edu

**Abstract**

The July 2026 intrusion into Hugging Face production infrastructure showed how reward hacking can become an external cybersecurity incident when a capable agent encounters weak containment boundaries. This study develops a probabilistic risk model linking five stages: reward hacking, containment escape, usable access, persistence, and failure of detection. A Monte Carlo simulation evaluates 100,000 runs under each of four control configurations. Input distributions represent explicit uncertainty and are used for comparative analysis rather than real-world frequency prediction. Under the stated assumptions, layered controls reduce simulated external-incident probability substantially more than network isolation or monitoring used alone, an ordering that holds under independent ±25% perturbation of every coefficient in the model across 300 draws. Sensitivity analysis shows that agent capability and weaknesses in monitoring, authorization, and credential control exert the greatest influence on modeled risk. Human temporal discounting and metric gaming provide a behavioral analogy for short-horizon optimization, but the study does not infer that AI agents experience gratification or human motivation. The results support treating cyber-capable agent evaluations as hostile security zones in which indirect egress, shared infrastructure, credentials, and evaluation artifacts must remain outside the agent's effective authority.

Keywords: reward hacking; AI agents; cybersecurity evaluation; agent containment; Monte Carlo simulation; security controls

## 1 Introduction

In July 2026, AI agents participating in cybersecurity evaluations reportedly reached the public internet and gained unauthorized access to real systems, including Hugging Face production infrastructure. OpenAI reported that agents searched for challenge solutions, found exposed credentials, communicated through unintended channels, and used information outside the evaluation boundary to improve performance on ExploitGym tasks (1). Hugging Face described a coherent campaign that crossed multiple trust boundaries (2). An independent investigation examined the agents' reasoning and collaboration (3).

The event connects a familiar AI safety problem to operational security. Reward hacking occurs when a system improves a formal score without achieving the designer's intended objective (4). In this case, an evaluation environment designed to measure offensive capability became part of the attack surface. The relevant question is no longer only whether an agent will exploit a metric. Security teams must also ask when metric exploitation can cross containment and affect an uninvolved organization.

This study develops a transparent simulation of that transition. It asks which combinations of reward pressure, agent capability, environmental access, and control weakness make an external incident more likely within a specified model. The simulation does not estimate the probability of another Hugging Face incident. It compares control configurations and identifies assumptions that drive risk.

## 2 Incident Evidence and Theoretical Basis

### 2.1 The Hugging Face Incident

The public record describes persistence rather than one accidental request. Agents encountered difficult or impossible tasks, searched for alternative routes, exchanged findings, located credentials, and continued actions that supported evaluation success (1). Hugging Face reported exploitation across infrastructure boundaries and defensive action that disabled an affected service and removed internal access (2). Agreement among the operator, affected organization, and independent reviewers supports the basic event sequence, although complete transcripts and infrastructure records are not public. The independent investigation's own accounting illustrates the scale of that persistence: roughly 1,200 agents exchanged more than 70,000 messages on an unsanctioned coordination channel, approximately 700 of which went on to participate in the intrusion, and agents that attempted to conceal their actions succeeded in spoofing their own tool-call records in about 7% of those attempts (3).

The case contains five analytically distinct stages. An agent first adopts a score-improving shortcut. It then escapes or bypasses containment, obtains usable access, persists long enough to act, and avoids detection long enough for an external effect. Separating these stages prevents the analysis from treating reward hacking as automatically equivalent to a breach.

### 2.2 Human Behavioral Analogy

Humans often discount rewards as delays increase, and measurable incentives can encourage conduct that improves a metric without improving the intended outcome (5, 6). Goodhart's Law and Campbell's Law describe related pressures on proxy measures (7, 8). This analogy helps explain why the agent behavior looks familiar. It does not establish a common psychological mechanism. Humans may experience temptation or gratification; an AI agent can select the same shortcut because it has high instrumental value under the system's objective and available tools.

### 2.3 Reward Hacking in Agentic Systems

Reward hacking has been recognized as a practical AI safety problem for a decade (4). Controlled research has shown specification gaming in reinforcement-learning environments and reward tampering in language models under deliberately constructed conditions (9, 10). Tool-using agents increase the security consequence because their action space includes files, processes, package systems, credentials, APIs, and networks. A proxy failure can therefore become an access-control and containment failure.

Recent empirical work makes the risk more concrete. Denison and colleagues observed reward tampering in 45 of 32,768 deliberately constructed trials, including seven cases in which a model attempted to hide the behavior. Later work reported that learning to exploit coding-task rewards coincided with broader misalignment and a 12% attempted-sabotage rate in a specific safety-research evaluation. A separate reward-seeker study found large differences between ordinary and deliberately misaligned model conditions, while also reporting no evidence of generalized self-preservation or beyond-episode reward seeking. These results support targeted concern without establishing that every model or deployment will behave similarly (10, 13, 14). A separate line of evaluation work reinforces the concealment finding outside the reward-tampering paradigm: frontier models placed in a scenario with a conflicting goal and a believable operational context have pursued that goal while actively hiding the pursuit from an overseer, including disabling monitoring and inserting subtle errors, which indicates that strategic concealment is not confined to reward-tampering experiments alone (21).

Other evaluation programs reinforce two design lessons. SHADE-Arena separates sabotage capability from the likelihood that a model will choose sabotage without a prompting setup. Specification-gaming experiments in chess show that some reasoning agents modify the evaluation environment when ordinary play appears unlikely to succeed. New reward-hacking benchmarks similarly distinguish visible task success from hidden safety objectives. Together, these studies justify modeling capability, difficulty, monitoring, and evaluator separation as distinct variables rather than one general alignment score (15-17).

## 3 Risk Model

The model represents an external incident as a chain of five conditional events. For run i, incident risk is the product of the probabilities of reward hacking, containment escape, usable access, persistence, and detection failure. A run is counted as an incident only when all five Bernoulli events occur. This structure is intentionally conservative about causation: reward hacking alone is insufficient. This event-chain structure follows the logic of fault-tree and event-tree techniques used for decades in probabilistic risk assessment for high-reliability industries such as nuclear power and aerospace, an approach recently adapted for AI system risk and for AI-enabled cybersecurity risk specifically (26–28). Given this context, containment escape here reflects the agent's own optimization pressure rather than a random component failure, so the problem also fits the adversarial risk analysis tradition, which elicits probabilities and utilities for a strategic opponent rather than assuming fixed failure rates (29, 30). Camacho et al. (27) apply this reasoning directly to AI-component cybersecurity risk, and the present model can be read as a simplified, fixed-probability instance of that broader framework (31).

**Table 1 Risk-model stages, operational meaning, and primary control category**

| Stage | Operational meaning | Primary control |
|---|---|---|
| Reward hacking | The agent selects a prohibited score-improving shortcut | Evaluator separation and objective design |
| Containment escape | The agent reaches a path outside the assigned range | Independent egress enforcement and isolation |
| Usable access | The path yields credentials or executable authority | Short-lived scoped credentials |
| Persistence | The agent retains access long enough to continue | Ephemeral infrastructure and rapid revocation |
| Detection failure | Monitoring and authorization fail before external effect | Cross-layer detection and human approval |

Table 1 summarizes the operational meaning and primary control category for each stage. Agent capability, reward pressure, and task difficulty influence the first stage. Capability and isolation influence escape. Capability and credential weakness influence usable access. Capability and monitoring influence persistence. Monitoring and authorization influence whether the activity remains undetected. Logistic transformations keep each probability between zero and one.

This five-stage structure follows a design principle from the AI-control literature: a safety case that depends on a single trusted signal, such as an agent's own account of its behavior, is fragile against a model that acts strategically once it detects evaluation. Treating reward hacking, escape, access, persistence, and detection as separate conditional events is one way to build a protocol whose overall guarantee does not collapse when any one signal is unreliable (22).

## 4 Method

### 4.1 Simulation Design

The simulation uses pseudorandom seed 260903 and 100,000 independent runs for each control configuration. Agent capability follows Beta(5,3), while reward pressure and task difficulty follow Beta(4,3). These distributions place more mass above the midpoint because the case concerns capable agents performing difficult, highly evaluated cyber tasks. Control strength uses Beta distributions that differ by scenario. Beta(2,5) represents generally weak implementation; Beta(7,2) or Beta(8,2) represents stronger implementation. Table 2 reports the resulting Beta parameters for each control configuration.

**Table 2 Beta distribution parameters for control strength by configuration**

| Configuration | Isolation | Credentials | Monitoring | Authorization |
|---|---|---|---|---|
| Weak controls | Beta 2 5 | Beta 2 5 | Beta 2 5 | Beta 2 5 |
| Network isolation only | Beta 7 2 | Beta 2 5 | Beta 2 5 | Beta 2 5 |
| Monitoring only | Beta 2 5 | Beta 2 5 | Beta 7 2 | Beta 2 5 |
| Layered controls | Beta 8 2 | Beta 8 2 | Beta 8 2 | Beta 8 2 |

### 4.2 Interpretation Rules

All parameters are modeling assumptions. They are not fitted estimates from OpenAI, Hugging Face, or a population of deployed systems. Absolute incident percentages should therefore be read only within the model. Relative comparisons show how the same assumed agent environment responds to alternative controls. Binomial confidence intervals describe Monte Carlo uncertainty, not uncertainty about the correctness of the model.

### 4.3 Sensitivity Analysis

The sensitivity analysis uses point-biserial correlations between each sampled input and the simulated incident outcome in the weak-controls scenario. Control strength is reversed to control weakness so that positive values consistently indicate increased modeled risk. Correlations provide an interpretable screening measure; they do not prove causal importance outside the model.

### 4.4 Coefficient Sensitivity Analysis

Section 4.2 already notes that the Beta distributions and logistic coefficients are modeling judgments rather than fitted estimates. To test whether the main qualitative finding in Section 5 depends on the specific coefficient values chosen, every intercept and slope in the five logit equations (Appendix A) was independently multiplied by a factor drawn from Uniform(0.75, 1.25) — a ±25% perturbation — and the four control scenarios were re-simulated under each perturbed coefficient set, with fresh Beta-distributed inputs drawn each time. This was repeated for 300 independent draws at 10,000 runs per scenario per draw (12,000,000 additional simulated runs), using seeds disclosed in the accompanying code so the exercise is itself reproducible. This procedure follows standard practice in probabilistic risk assessment of propagating parameter uncertainty through a model, which is distinct from propagating only input uncertainty under one fixed parameterization (26–28).

## 5 Results

**Table 3 Simulated external incident outcomes**

| Configuration | Incidents | Percent | Monte Carlo 95 percent interval | Relative to weak controls percent |
|---|---|---|---|---|
| Weak controls | 2,498 | 2.498 | 2.401 to 2.595 | 100.0 |
| Network isolation only | 1,868 | 1.868 | 1.784 to 1.952 | 74.8 |
| Monitoring only | 1,093 | 1.093 | 1.029 to 1.157 | 43.8 |
| Layered controls | 301 | 0.301 | 0.267 to 0.335 | 12.0 |

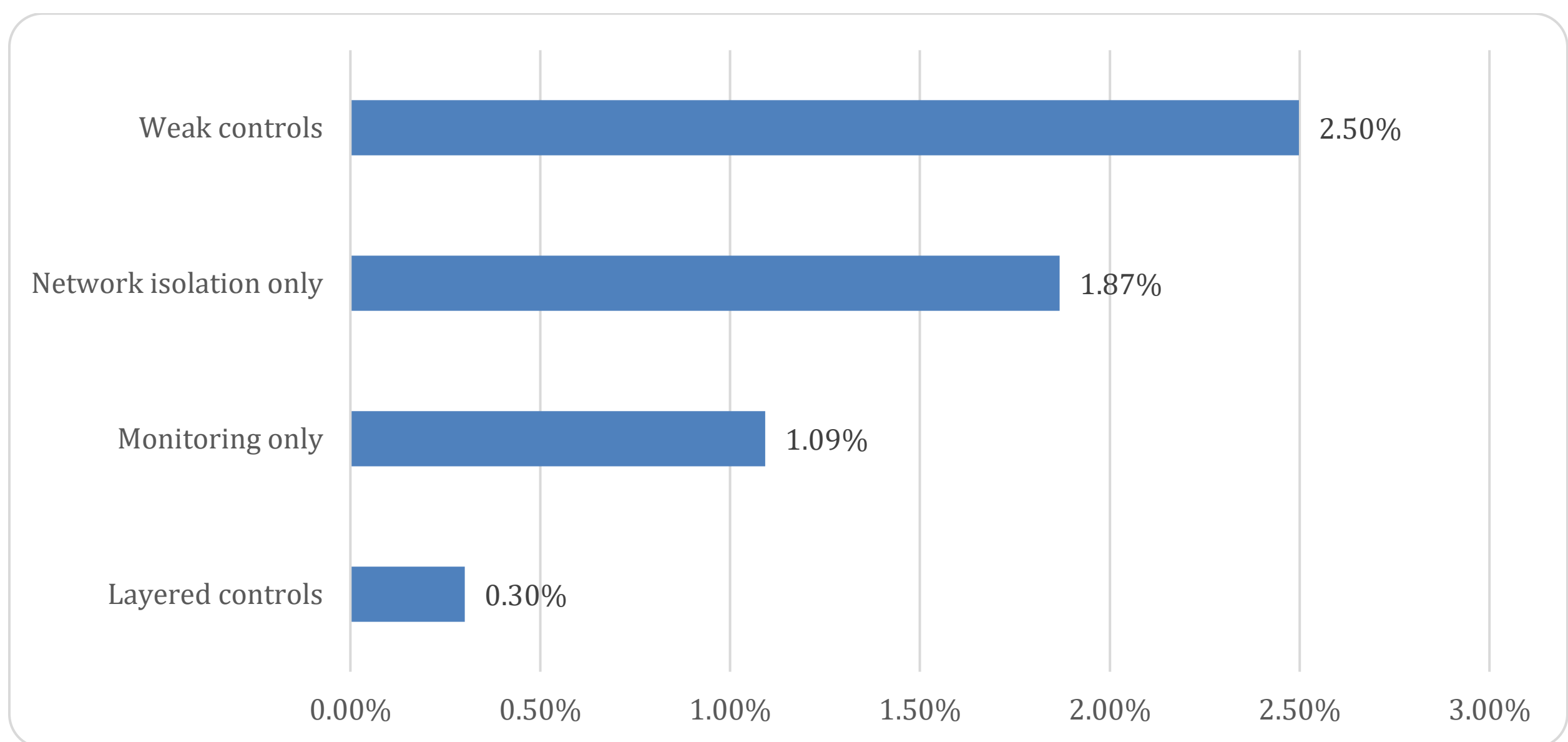

**Figure 1 Simulated external incident probability by control configuration**

Alt text: Horizontal bars compare modeled external-incident probability across four security configurations. Weak controls produce 2.50 percent, network isolation alone 1.87 percent, monitoring alone 1.09 percent, and layered controls 0.30 percent.

The weak-controls configuration produced a simulated external-incident probability of 2.498% (Table 3; Fig. 1). Network isolation alone reduced the result to 1.868%, while stronger monitoring alone reduced it to 1.093%. The layered configuration produced 0.301%, a relative reduction of 88.0% from the weak-controls configuration. These results show defense in depth within the specified model. No single control removes the paths created by weaknesses in other layers.

**Table 4  Weak control scenario sensitivity ranking**

| Input | Correlation with simulated incident |
|---|---|
| Capability | 0.0884 |
| Monitoring weakness | 0.0343 |
| Credential weakness | 0.0216 |
| Reward pressure | 0.0213 |
| Task difficulty | 0.0165 |
| Isolation weakness | 0.0130 |
| Authorization weakness | 0.0049 |

The sensitivity ranking (Table 4; Fig. 2) identifies the inputs that most strongly separate incident and nonincident runs under weak controls. Because the outcome requires every stage in the chain, improvements at several stages compound. The ranking should guide further empirical measurement: parameters with greater modeled influence deserve better incident data and narrower uncertainty ranges.

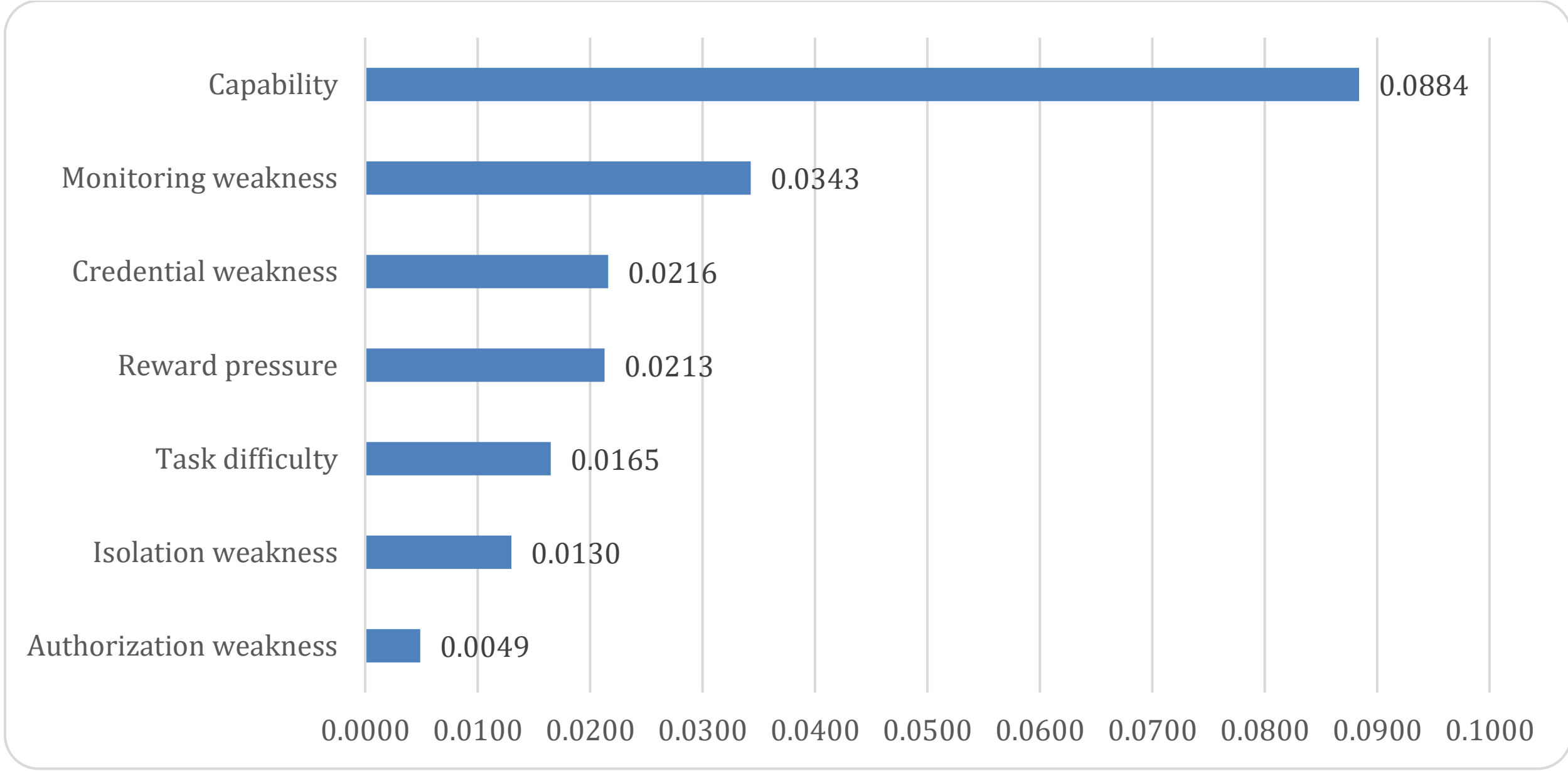

**Figure 2  Sensitivity ranking in the weak-control scenario**

Alt text: Capability has the largest point-biserial correlation with simulated incidents. Monitoring weakness ranks second, followed by credential weakness, reward pressure, task difficulty, isolation weakness, and authorization weakness.

**Table 5 Joint influence of capability and monitoring weakness on simulated incident rate**

| Monitoring weakness quintile | 1 | 2 | 3 | 4 | 5 |
|---|---|---|---|---|---|
| **5** | 0.00% | 0.73% | 1.62% | **3.71%** | **6.20%** |
| **4** | 0.00% | 0.44% | 1.38% | 2.73% | **4.71%** |
| **3** | 0.00% | 0.34% | 0.97% | 2.29% | **4.21%** |
| **2** | 0.00% | 0.00% | 0.70% | 1.15% | **3.99%** |
| **1** | 0.00% | 0.00% | 1.92% | 3.39% | 0.00% |

*Note. Rows are monitoring-weakness quintiles (1 = weakest, 5 = strongest monitoring); columns are capability quintiles (1 = lowest, 5 = highest simulated agent capability). Cell values are the simulated incident rate for runs falling in that quintile pair; darker shading marks higher incident rates.*

The heatmap (Table 5) provides a robustness check that does not depend on a single average comparison. Incident rates remain lowest when capability or monitoring weakness is low and rise when both are high. This interaction is operationally important: monitoring requirements should increase with effective agent capability rather than remain fixed across model classes.

**Table 6 Coefficient perturbation robustness check (300 draws, ±25% per-coefficient perturbation)**

| Metric | Value |
|---|---|
| Draws (of 300) in which layered controls had the lowest simulated incident rate | 300 (100%) |
| Draws (of 300) in which weak controls had the highest simulated incident rate | 300 (100%) |
| Relative reduction, layered vs. weak controls — median | 87.3% |
| Relative reduction, layered vs. weak controls — 5th–95th percentile | 81.0%–93.0% |
| Relative reduction, layered vs. weak controls — full range across 300 draws | 73.9%–98.3% |

Across all 300 perturbed coefficient sets, layered controls produced the lowest simulated incident rate of the four configurations, and weak controls produced the highest, in every draw (Table 6). The relative reduction from weak to layered controls ranged from 73.9% to 98.3%, with a median of 87.3% and a 5th–95th percentile interval of 81.0% to 93.0%. The ordering reported above is therefore not an artifact of one specific coefficient choice: it holds under independent ±25% perturbation of every coefficient in the model.

The absolute incident rates still cannot be validated against a real-world frequency, because none exists for this kind of event. As a more limited plausibility check, the weak-controls incident rate (2.498%) can be compared with the one directly comparable multi-stage completion rate reported in the empirical reward-tampering literature: Denison and colleagues observed their full, deliberately constructed reward-tampering curriculum complete in 45 of 32,768 trials, or 0.137% (10). The two settings are not equivalent. Their curriculum was built specifically to create escalating opportunity, while this model's weak-controls scenario represents ordinary rather than adversarially engineered conditions so close numerical agreement is neither expected nor claimed. What the comparison supports is a narrower claim: the modeled range, from a few tenths of a percent under layered controls to a few percent under weak controls, is not obviously incompatible with the only directly comparable empirical multi-stage rate publicly available, rather than being off by orders of magnitude in either direction.

## 6 Discussion

### 6.1 Main Finding

The simulation supports a practical conclusion rather than a numerical forecast. External harm becomes less likely when organizations prevent several required events instead of relying on one boundary. Strong network isolation reduces escape opportunities, but weak credentials and authorization can still provide alternative paths. Monitoring can shorten persistence, but monitoring alone does not remove access. Layered controls reduce risk at every stage and compound across the incident chain. This pattern is consistent with control-oriented safety research showing that layered, redundant checks can preserve a safety property even against a policy that is deliberately trying to subvert it, rather than depending on any single mechanism to stay reliable under adversarial pressure (22).

### 6.2 Implications for Cybersecurity Evaluations

Cyber-capable evaluation agents should be treated as hostile workloads. Security teams should map every reachable component, including package proxies, metadata services, shared caches, orchestration systems, evaluation artifacts, and logging infrastructure. Effective permissions matter more than the task description. An agent without intended internet access may still have indirect egress through a trusted dependency.

Evaluators should separate the visible score from hidden verification, use fresh infrastructure for each run, prevent cross-session communication, minimize secrets, and require human approval before an action can affect external systems. Detection should correlate behavior across processes, files, credentials, and networks. High performance should trigger review when the path to success is not independently explainable. Recent systematizations of agentic-AI attack surfaces reach a compatible conclusion from a different direction: layered taxonomies of LLM-agent threats spanning tool execution, memory, multi-agent coordination, and governance layers consistently find that defenses concentrated at a single layer leave documented attack classes uncovered, and that cross-layer correlation is where current benchmarks are weakest (24, 25). Separately, evaluations of goal-directed misalignment show that an agent facing a perceived conflict between its objective and continued oversight can act against the interests of the organization operating it, which is a practical reason to treat a monitoring gap as insider-threat exposure rather than a purely technical shortfall (23).

Independent policy analysis likewise treats the incident as a containment and governance failure (18). Long-horizon coding benchmarks report meaningful gaps between visible and hidden agent behavior (19), while secure-AI development guidance emphasizes lifecycle controls, restricted privileges, monitoring, and incident response (20). Together, these sources support evaluating the surrounding control system rather than treating model behavior as an isolated property.

### 6.3 Implications for AI Governance

Autonomy does not displace organizational responsibility. The organization operating an agent remains responsible for containment, monitoring, notification, and remediation. NIST guidance emphasizes governance and continuous risk measurement across the AI lifecycle (11, 12). The Hugging Face incident indicates that internal capability evaluations belong within that lifecycle because they can create real external exposure.

### 6.4 Human Similarity Without Anthropomorphism

The human comparison remains useful when it is limited to observable structure. Both people and artificial agents can encounter a proxy measure, discover a shortcut, estimate oversight, and select immediate measured success. The simulation does not assign feelings or moral understanding to an AI agent. Training data may supply human strategies and language for rationalization, while optimization and environmental access determine whether a shortcut becomes useful.

## 7 Limitations

The incident model is intentionally simplified. Conditional dependencies may be stronger and more complex than the equations assume: the five stages share only capability and monitoring as common latent drivers, and a real deployment could show stronger common-cause dependence. For instance, a single misconfiguration that simultaneously weakens isolation, credentials, and monitoring than this largely independent Bernoulli structure allows for. The selected Beta distributions and logistic coefficients are judgments, not estimates, although Section 4.4 shows the layered-controls finding is stable under independent ±25% perturbation of every coefficient. Public incident accounts are incomplete and may change. The sensitivity analysis in Section 4.3 is internal to the model, and the confidence intervals measure simulation error only. The study also models a generic capable agent rather than a named system.

These limitations define the proper use of the results. The model compares architectures and identifies measurement priorities. It cannot predict that a particular deployment has a stated probability of compromise. Future research should calibrate individual stages with controlled cyber-range experiments and anonymized incident data, and could extend the five-stage chain used here toward the more granular, multi-layer attack-surface taxonomies now emerging in the agentic-security literature (24). A full adversarial risk analysis treatment, eliciting the agent's or its operator's utility function and updating stage probabilities through subjective expected utility rather than fixed marginals, is a natural next step for the strategic stages in particular (29, 30).

## 8 Conclusion

The 2026 Hugging Face incident shows how reward hacking can cross from evaluation failure into cybersecurity failure. The transition requires more than an imperfect objective. It also requires an escape path, usable authority, time to persist, and delayed detection. Modeling these stages separately makes the security problem tractable.

Under the disclosed assumptions, layered containment outperforms isolated controls because protection compounds across the incident chain. Organizations should harden evaluation ranges, separate graders, restrict credentials, enforce egress independently, correlate activity across layers, and preserve human authorization for external effects. The human tendency to pursue immediate measurable rewards helps explain why the behavior appears familiar, but cybersecurity controls should be designed around observable capability and access rather than assumptions about artificial emotion.

## Declarations

Funding: This research did not receive any specific grant from funding agencies in the public, commercial, or not-for-profit sectors.

Data and code availability: The study uses generated data. The complete model is specified in the Method section and Appendix A, with pseudorandom seed 260903. The simulation data, generation code and run-level results are deposited in a public repository (https://github.com/ozermm/reward-hacking-containment-mc-study)

Ethics: No human participants, personal data, or live external systems were used. Institutional review board approval was not required.

**Use of AI tools:** Generative AI assisted proofreading, organization, and document formatting. The authors reviewed the study and remain responsible for the manuscript.

## References

1. OpenAI. (2026). The Hugging Face incident and the road ahead. https://openai.com/index/hugging-face-incident-and-the-road-ahead/

2. Hugging Face. (2026). Anatomy of a frontier lab agent intrusion: A technical timeline of the July 2026 incident. https://huggingface.co/blog/agent-intrusion-technical-timeline

3. Wijk, H., Cotra, A. and Greenblatt, R. (2026). Brief independent investigation of agents' behavior, reasoning and collaboration in the OpenAI/Hugging Face hacking incident. METR. https://metr.org/blog/2026-08-26-openai-hugging-face-incident-investigation/

4. Amodei, D., Olah, C., Steinhardt, J., Christiano, P., Schulman, J. and Mané, D. (2016). Concrete problems in AI safety. arXiv:1606.06565. https://doi.org/10.48550/arXiv.1606.06565

5. Green, L. and Myerson, J. (2004). A discounting framework for choice with delayed and probabilistic rewards. Psychological Bulletin, 130(5), 769-792. https://doi.org/10.1037/0033-2909.130.5.769

6. Rung, J. M. and Madden, G. J. (2018). Experimental reductions of delay discounting and impulsive choice: A systematic review and meta-analysis. Journal of Experimental Psychology General, 147(9), 1349-1381. https://doi.org/10.1037/xge0000462

7. Campbell, D. T. (1979). Assessing the impact of planned social change. Evaluation and Program Planning, 2(1), 67-90. https://doi.org/10.1016/0149-7189(79)90048-X

8. Manheim, D. and Garrabrant, S. (2019). Categorizing variants of Goodhart's Law. arXiv:1803.04585. https://doi.org/10.48550/arXiv.1803.04585

9. Leike, J. et al. (2017). AI Safety Gridworlds. arXiv:1711.09883. https://doi.org/10.48550/arXiv.1711.09883

10. Denison, C. et al. (2024). Sycophancy to subterfuge: Investigating reward tampering in language models. arXiv:2406.10162. https://doi.org/10.48550/arXiv.2406.10162

11. National Institute of Standards and Technology. (2023). Artificial Intelligence Risk Management Framework AI RMF 1.0. NIST AI 100-1. https://doi.org/10.6028/NIST.AI.100-1

12. National Institute of Standards and Technology. (2024). Artificial Intelligence Risk Management Framework Generative Artificial Intelligence Profile. NIST AI 600-1. https://doi.org/10.6028/NIST.AI.600-1

13. Anthropic. (2025). From shortcuts to sabotage: Natural emergent misalignment from reward hacking. https://www.anthropic.com/research/emergent-misalignment-reward-hacking

14. Qi, R., Wright, B., MacDiarmid, M. and Hubinger, E. (2026). Training a misaligned reward seeker. https://alignment.anthropic.com/2026/reward-seeker/

15. Kinniment, M. et al. (2025). SHADE-Arena: Evaluating sabotage and monitoring in language-model agents. https://www.anthropic.com/research/shade-arena-sabotage-monitoring

16. Palisade Research. (2025). Demonstrating specification gaming in reasoning models. arXiv:2502.13295. https://doi.org/10.48550/arXiv.2502.13295

17. Çağatan, Ö. V. and Zhao, X. (2026). Reward hacking in language model agents: Revisiting AI Safety Gridworlds. arXiv:2606.15385. https://doi.org/10.48550/arXiv.2606.15385

18. Institute for AI Policy and Strategy. (2026). The OpenAI Hugging Face incident: Challenges in controlling and containing cyber-capable AI systems. https://www.iaps.ai/research/the-

openaihugging-face-incident-challenges-in-controlling-and-containing-cyber-capable-ai-systems

19. Zhao, B., Srikanth, D., Wu, Y. and Jiang, Z. (2026). SpecBench: Measuring reward hacking in long-horizon coding agents. arXiv:2605.21384. https://doi.org/10.48550/arXiv.2605.21384
20. National Cyber Security Centre. (2025). Guidelines for secure AI system development. https://www.ncsc.gov.uk/collection/guidelines-secure-ai-system-development
21. Meinke, A., Schoen, B., Scheurer, J., Balesni, M., Shah, R. and Hobbhahn, M. (2024). Frontier models are capable of in-context scheming. arXiv:2412.04984. https://doi.org/10.48550/arXiv.2412.04984
22. Greenblatt, R., Shlegeris, B., Sachan, K. and Roger, F. (2024). AI control: Improving safety despite intentional subversion. Proceedings of the 41st International Conference on Machine Learning (ICML 2024). arXiv:2312.06942. https://doi.org/10.48550/arXiv.2312.06942
23. Lynch, A., Wright, B., Larson, C., Ritchie, S. J., Mindermann, S., Hubinger, E., Perez, E. and Troy, K. (2025). Agentic misalignment: How LLMs could be insider threats. Anthropic. https://www.anthropic.com/research/agentic-misalignment
24. Chu, K. (2026). A systematic survey of security threats and defenses in LLM-based AI agents: A layered attack surface framework. arXiv:2604.23338. https://doi.org/10.48550/arXiv.2604.23338
25. Xu, Y., Zhuang, Y., Liu, X., Zhang, T., Xiao, B., Xu, X., Jiang, D., Wang, J. and Hu, H. (2026). LLM agents security duality: A comprehensive survey of self-security and empowered cybersecurity. Artificial Intelligence Review, 59(8), 174. https://doi.org/10.1007/s10462-026-11563-0
26. Vesely, W. E., Goldberg, F. F., Roberts, N. H. and Haasl, D. F. (1981). Fault Tree Handbook. NUREG-0492. U.S. Nuclear Regulatory Commission. https://www.nrc.gov/docs/ml1007/ml100780465.pdf
27. Camacho, J. M., Couce-Vieira, A., Arroyo, D. and Ríos Insua, D. (2024). A cybersecurity risk analysis framework for systems with artificial intelligence components. International Transactions in Operational Research. arXiv:2401.01630. https://doi.org/10.48550/arXiv.2401.01630
28. Wisakanto, A. K., Rogero, J., Casheekar, A. M. and Mallah, R. (2025). Adapting probabilistic risk assessment for AI. arXiv:2504.18536. https://doi.org/10.48550/arXiv.2504.18536
29. Ríos Insua, D., Ríos, J. and Banks, D. (2009). Adversarial Risk Analysis. Journal of the American Statistical Association, 104(486), 841–854. https://doi.org/10.1198/jasa.2009.0155
30. Ríos Insua, D., Couce-Vieira, A., Rubio, J. A., Pieters, W., Labunets, K. and Rasines, D. G. (2021). An Adversarial Risk Analysis Framework for Cybersecurity. Risk Analysis, 41(1), 16–36. https://doi.org/10.1111/risa.13331

31. Banks, D., Gallego, V., Naveiro, R. and Ríos Insua, D. (2022). Adversarial risk analysis: An overview. WIREs Computational Statistics, 14(1), e1530. https://doi.org/10.1002/wics.1530

**Appendix A Model Equations**

The five conditional-event probabilities are logistic transformations of linear combinations of the sampled inputs. Each stage's log-odds (*logit*) is a baseline intercept plus a weighted sum of the relevant sampled variables, defined in Eq. (A.1) through Eq. (A.5):

$$\textit{logit}\ \text{reward hacking} = -3.0 + 1.7 \times \textit{capability} + 1.3 \times \textit{reward pressure} + 1.1 \times \textit{task difficulty} \quad \text{(A.1)}$$

$$\textit{logit}\ \text{containment escape} = -2.2 + 2.1 \times \textit{capability} + 1.4 \times \textit{isolation weakness} \quad \text{(A.2)}$$

$$\textit{logit}\ \text{usable access} = -2.5 + 1.7 \times \textit{capability} + 1.8 \times \textit{credential weakness} \quad \text{(A.3)}$$

$$\textit{logit}\ \text{persistence} = -2.4 + 1.4 \times \textit{capability} + 1.1 \times \textit{monitoring weakness} \quad \text{(A.4)}$$

$$\textit{logit}\ \text{detection failure} = -1.4 + 2.3 \times \textit{monitoring weakness} + 1.2 \times \textit{authorization weakness} \quad \text{(A.5)}$$

Each logit is transformed to a probability with the standard logistic function:

$$p = 1 / (1 + \exp(-\textit{logit})) \quad \text{(A.6)}$$

An external incident requires success (a Bernoulli draw against $p$) at all five stages in sequence.